\documentclass{bmvc2k}
\usepackage{amsfonts}
\usepackage{amsmath}
\usepackage{color}
\usepackage{multirow}
\usepackage{wrapfig}
\usepackage{colortbl}
\usepackage{enumitem}
\usepackage{comment}
\usepackage{caption}
\usepackage{graphicx}
\usepackage{tabularx}
\usepackage{booktabs}
\usepackage{subcaption}
\usepackage{xspace}
\usepackage{microtype}
\usepackage{pifont}
\newcommand{\cmark}{\textcolor{black}{\ding{51}}} % checkmark symbol
\newcommand{\xmark}{\textcolor{black}{\ding{55}}}   % crossmark symbol

\def\onedot{.\@\xspace}
 
\def\ie{\emph{i.e}\onedot}

\def\etal{\emph{et al}\onedot}

\newcommand{\Sref}[1]{Sec.~\ref{#1}}
\newcommand{\Eref}[1]{Eq.~(\ref{#1})}
\newcommand{\Fref}[1]{Fig.~\ref{#1}}
\newcommand{\Tref}[1]{Table~\ref{#1}}

\newcommand{\bx}{{\mathbf{x}}}
\newcommand{\by}{{\mathbf{y}}}
\newcommand{\bz}{{\mathbf{z}}}

\newcommand{\bA}{\mathbf{A}}

\newcommand{\bW}{\mathbf{W}}

\newcommand{\calL}{{\mathcal{L}}}

\newcommand{\calN}{{\mathcal{N}}}

\newcommand{\bmu}{\mbox{\boldmath $\mu$}}

\newcommand{\bsigma}{\mbox{\boldmath $\sigma$}}

\newcommand{\bSigma}{\mbox{\boldmath $\Sigma$}}

\newcommand{\bPsi}{\mbox{\boldmath $\Psi$}}

\newcommand{\Real}{\mathbb R}

\newcommand{\be}{\begin{eqnarray}}
\newcommand{\ee}{\end{eqnarray}}
\newcommand{\bee}{\begin{eqnarray*}}
\newcommand{\eee}{\end{eqnarray*}}

\newcommand{\matrixb}{\left[ \begin{array}}
\newcommand{\matrixe}{\end{array} \right]}

\title{Learning Correlation-aware Aleatoric Uncertainty for 3D Hand Pose Estimation}

\addauthor{Lee Chae-Yeon}{chaeyeon.lee@postech.ac.kr}{1}
\addauthor{Nam Hyeon-Woo}{hyeonw.nam@postech.ac.kr}{2}
\addauthor{Tae-Hyun Oh}{thoh.kaist.ac.kr@gmail.com}{3}

\addinstitution{
 Grad. School of AI\\
 POSTECH, South Korea
}
\addinstitution{
 Dept. of Electrical Engineering\\
 POSTECH, South Korea
}
\addinstitution{
School of Computing\\
KAIST, South Korea
}

\runninghead{Chae-Yeon ET AL.}{Learning Correlation-aware Aleatoric Uncertainty}

\def\etal{\emph{et al}\bmvaOneDot}

\begin{document}

\maketitle
\begin{abstract}
3D hand pose estimation is a fundamental task in understanding human hands. However, accurately estimating 3D hand poses remains challenging due to the complex movement of hands, self-similarity, and frequent occlusions. In this work, we address two limitations: the inability of existing 3D hand pose estimation methods to estimate aleatoric (data) uncertainty, and the lack of uncertainty modeling that incorporates joint correlation knowledge, which has not been thoroughly investigated. To this end, we introduce aleatoric uncertainty modeling into the 3D hand pose estimation framework, aiming to achieve a better trade-off between modeling joint correlations and computational efficiency. We propose a novel parameterization that leverages a single linear layer to capture intrinsic correlations among hand joints. This is enabled by formulating the hand joint output space as a probabilistic distribution, allowing the linear layer to capture joint correlations.
%in the output.
Our proposed parameterization is used as a task head layer, and can be applied as an add-on module on top of the existing models.
Our experiments demonstrate that our parameterization for uncertainty modeling outperforms existing approaches. Furthermore, the 3D hand pose estimation model equipped with our uncertainty head achieves favorable accuracy in 3D hand pose estimation while introducing new uncertainty modeling capability to the model.
% significantly outperforms existing approaches in aleatoric uncertainty estimation, while maintaining accurate 3D hand pose estimation performance.
The project page is available at \href{https://hand-uncertainty.github.io/}{https://hand-uncertainty.github.io/}.
\end{abstract}    
\section{Introduction}
\label{sec:intro}

Understanding human hands is fundamental for applications ranging from robotics to AR/VR~\cite{handa2020dexpilot, chi2024estimating}. 
The ability to perceive and interpret human hands significantly enhances the dexterity of robot-assisted tasks and ensures seamless human-like robot-object interactions.
In learning from human demonstrations, robots can acquire human hand behavior by observing these demonstrations~\cite{handa2020dexpilot,du2022multi,qin2022dexmv}.
In this work, we address two key limitations found in state-of-the-art methods for estimating 3D hand pose.

\emph{(1) Inability to estimate the aleatoric (data) uncertainty.} While recent works in learning-based 3D hand pose estimation~\cite{li2022interacting,yu2023acr,zhang2021interacting,meng20223d,kim2021end,yang2022artiboost,hasson2019learning,zuo2023reconstructing,pavlakos2024reconstructing} have made significant progress, accurate 3D hand-pose estimation inherently faces uncertainty from in-the-wild video.
These uncertainty arise from the high number of degrees of freedom present in hands~\cite{bilal2011vision}, frequent occurrence of self-similarity and occlusion~\cite{moon2020interhand2,smith2020constraining}, and motion blur due to their dynamic nature~\cite{park20243d,oh2023recovering}. 
Against these observation noises, 
% The hand estimation models 
% network
% should be able to 
quantifying aleatoric uncertainty enhances the confidence of the hand estimation models
% that is sourced from data noise, in order 
to be deployed in real-world applications. 
%This is particularly important in decision-making for high-risk tasks 
%For example, robots need to know what they do not know or understand the confidence level in their decision to avoid mistakes. 
%This capability enables systems to make informed decisions and ensure robust performance in the presence of ambiguity.
\emph{(2) Uncertainty modeling in the absence of joint correlation knowledge.} Human hand offers many individual degrees of freedom, yet joint movements are correlated~\cite{hager2000quantifying,schieber1995muscular,ingram2008statistics,liu2021quantitative,liu2016analysis}. Although the uncertainty of one hand joint can influence the uncertainty of another joint, previous works~\cite{uncertainty, oh2018modeling} model uncertainty entry-wise independently under the independent assumption due to computational and parameter efficiency.

In this work, we introduce aleatoric uncertainty modeling into the 3D hand pose estimation framework, achieving a better trade-off between correlation modeling and efficiency. We propose a novel parametrization that leverages a single linear layer to capture the intrinsic correlations among hand joints. To enable this, we design a probabilistic hand joint output space that facilitates uncertainty modeling with consideration of joint dependencies. Specifically, we begin by training a pre-trained large model~\cite{pavlakos2024reconstructing} for aleatoric uncertainty estimation by adding a transformer head that regresses the per-joint variance. We use a Gaussian negative log-likelihood 
% (NLL)
loss with a diagonal covariance matrix, which enables the network to predict variances that represent the uncertainty of hand joints under the independence assumption. The estimated uncertainty defines a probabilistic output space, from which we draw samples and feed them into a single linear layer to transform the output space into a correlation-aware space. Our parametrization serves as a mid-representation between diagonal and full covariance matrix parametrizations. It provides higher expressiveness for capturing joint correlations than the diagonal form, yet requires significantly fewer parameters than the full covariance parametrization.

We demonstrate the effectiveness of our method on two standard benchmarks for 3D hand pose estimation, FreiHAND~\cite{zimmermann2019freihand} and HO3Dv2~\cite{hampali2020honnotate}. 
Our method outperforms
% achieves better performance 
% on uncertainty estimation compared to 
existing aleatoric uncertainty modeling methods on uncertainty estimation. The key to our method's effectiveness is the formulation of the hand joint output space as a probabilistic distribution, which enables the linear layer to effectively learn hand joint correlations. This approach allows for an analytic representation of a structured covariance matrix, facilitating direct estimation of uncertainty. Furthermore, our method maintains competitive accuracy in 3D hand pose estimation, demonstrating that modeling uncertainty does not compromise pose estimation performance.
Our main contributions are summarized as follows:
\begin{itemize}[itemsep=0pt]
    \item We introduce aleatoric uncertainty modeling into 3D hand pose estimation framework.
    \item We propose a novel parameterization that leverages a single linear layer to effectively model inherent hand joint correlations, which is enabled by formulating the hand joint output space as a probabilistic distribution. 
    \item Comprehensive experiments demonstrate that our proposed method significantly outperforms existing aleatoric uncertainty modeling methods in uncertainty estimation, while maintaining accurate 3D hand pose estimation performance.
\end{itemize}

\section{Related Work}
\label{sec:related_work}
\noindent\textbf{3D hand pose estimation.}
3D hand pose estimation from a single RGB image has received a great attention for understanding complex hand interaction. The advent of deep learning has especially improved the performance of 3D hand pose estimation compared to hand-crafted geometric features. Most existing works~\cite{fan2023arctic, zhang2021interacting,baek2019pushing,rong2021frankmocap,pavlakos2024reconstructing} leverage the MANO parametric hand model~\cite{romero2022embodied} and regress the hand pose and shape parameters directly from an RGB image. Other works~\cite{choi2020pose2mesh,li2022interacting,ge20193d,kulon2020weakly} follow a non-parametric approach and regress mesh vertex coordinates directly from images for a more fine-grained reconstruction of hand surfaces.
Another line of work~\cite{moon2020interhand2, kim2021end, cai2018weakly, meng20223d} infer 3D hand joint positions of $x, y$ and $z$, which serve as a skeletal representation of hand posture.
More recently, HaMeR~\cite{pavlakos2024reconstructing} exploits 
% have shifted from convolutional neural networks to 
transformer networks~\cite{vaswani2017attention} and train on a large dataset, achieving robust 3D hand reconstruction and strong generalization to in-the-wild images. 
% These researches have only focused on the accurate prediction of hand parameters, but do not deal with the uncertainty.
% In this work, we 
Our work addresses aleatoric uncertainty in 3D hand pose estimation, which has been underexplored.
% in prior literature. 

\noindent\textbf{Uncertainty in deep learning.}
Uncertainty in deep learning can be categorized into aleatoric and epistemic uncertainties~\cite{der2009aleatory}.
Aleatoric uncertainty is attributed to the non-deterministic nature known as data uncertainty.
Epistemic uncertainty arises from model uncertainty due to insufficient knowledge learned
from the training data. In this work, we focus on aleatoric uncertainty in 3D hand joint positions. We assume
that the uncertainty is heteroscedastic~\cite{kendall2017uncertainties}, indicating that it depends on the inputs to the model, as certain inputs may inherently have higher uncertainty than others. A commonly used approach is to estimate the probability distribution over the output, and train the network by minimizing the negative log-likelihood (NLL) of the ground truth~\cite{kendall2017uncertainties,zhang2024heteroscedastic,bramlage2023plausible}. 
Caramalau~\etal~\cite{caramalau2021active} demonstrates that jointly modeling aleatoric and epistemic uncertainties is effective when applying an active learning framework to estimating 3D hand pose from a single depth image.
However, existing approaches for estimating 3D hand pose from a single RGB image either overlook the heteroscedastic uncertainty in input images or rely on latent distributions to implicitly capture aleatoric uncertainty through sampling~\cite{wang2022handflow,zuo2023reconstructing}, and they are not publicly available. Zhang~\etal~\cite{zhang2024weakly} applies the NLL loss for 3D hand reconstruction to model output-space uncertainty; however, their method is limited to modeling uncertainty in 2D hand joint positions, which may not accurately reflect the uncertainty in 3D joint space. AMVUR~\cite{jiang2023probabilistic} adopts a probabilistic framework for 3D hand pose and shape estimation, inherently supporting uncertainty quantification in 3D hand joint positions.
In this work, we explicitly model joint-wise uncertainty in the output space by learning a Gaussian distribution over 3D hand joint positions and incorporating it into the training objective through a simple yet effective NLL loss. Furthermore, we model inter-joint dependencies using a single linear transformation, enabling an analytic formulation of the structured uncertainty without relying on sampling-based approximations.
%To the best of our knowledge, this is the first approach to comprehensively model aleatoric uncertainty in 3D hand joint positions using negative log-likelihood (NLL).
\section{Method}
\label{sec:method}

\subsection{Preliminary}
\label{sec:preliminary}
% Most of recent deep neural network based approaches
% % The recent works on 3D hand pose estimation exploits the deep neural networks~
% \cite{li2022interacting,yu2023acr,zhang2021interacting,meng20223d,kim2021end,yang2022artiboost,hasson2019learning,zuo2023reconstructing,pavlakos2024reconstructing}
% % , which 
% estimate 3D hands pose directly. However, hands inherently possess fast and complex motions with similar appearances between two hands. This introduces ambiguities such as blurring and self-occlusion between hands.
% Uncertainty modeling enables models to quantify the uncertainty of their decisions, 
% % and the demand for uncertainty modeling is required 
% demanding
% in 3D hand pose estimation.
% In uncertainty modeling, two primary types are aleatoric and epistemic uncertainties. 
% In this paper, we focus on the aleatoric uncertainty. 

\begin{figure*}
\begin{center}
\includegraphics[width=\textwidth, keepaspectratio]{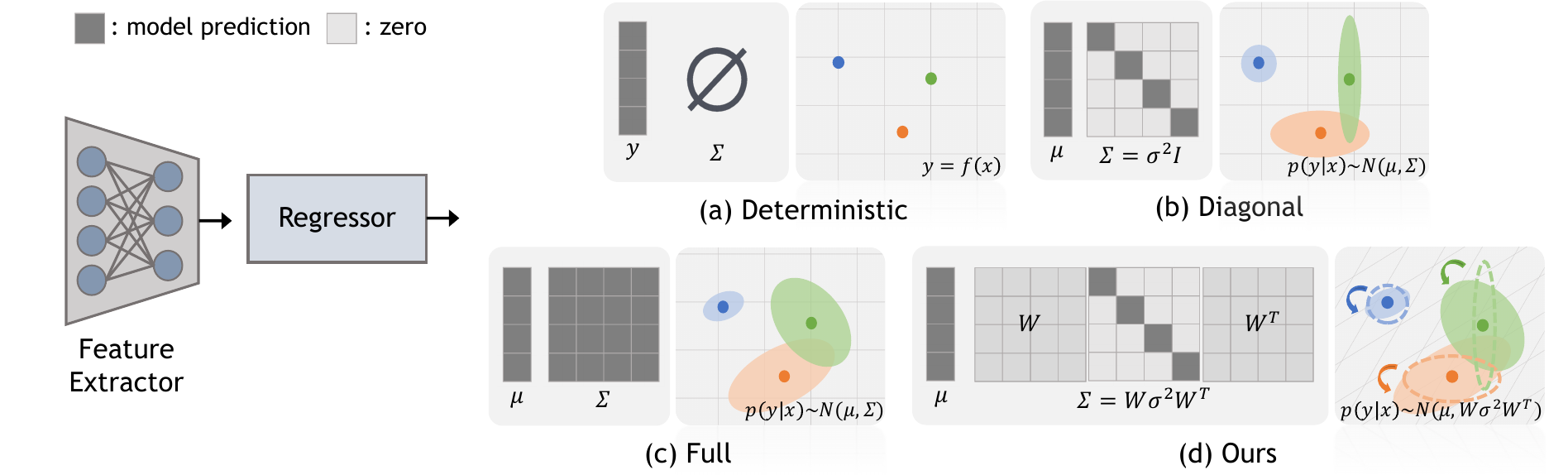}
\end{center}
   \caption{\textbf{Illustration of deterministic modeling and correlation modelings of uncertainty.} (a) Deterministic modeling produces a single deterministic output, represented as point embedding in the output space; (b) Diagonal and (c) Full correlation modeling of output produce means and covariances of a Gaussian distribution, where the uncertainty of the prediction is modeled by its variance. (d) Ours learns the variation of each dimension in the output space and shared weight $\bW$ which captures intrinsic hand joint dependencies, so that our covariance matrix is represented by $\bSigma=\bW\bsigma^2\bW^\top$.}  
\label{fig:teaser}
\end{figure*}

\noindent\textbf{Notation.}
We denote $\bx \in \Real^{d_{i}}$ as the input image, $f: \Real^{d_{i}} \rightarrow \Real^{d_{f}}$ as the feature extractor, $g: \Real^{d_{f}} \rightarrow \Real^{d_{o}}$ as the parameter regressor, and $\mathop{\mathrm{diag}(\cdot)}$ outputs a diagonal matrix given a vector.
% , where the diagonal is filled with the given vector.
As shown in \Fref{fig:teaser}a, the model without uncertainty modeling is denoted as $g(f(\bx))$.
The uncertainty modeling~\cite{uncertainty} modifies the regressor $g$ such that $g(f(\bx)) \sim \calN(\bmu, \bSigma)$ where $\calN$ is the Gaussian distribution with a mean $\bmu \in \Real^{d_{o}}$
% a mean vector,
and a covariance $\bSigma \in \Real^{d_{o} \times d_{o}}$.
% a covariance matrix.
The subsequent paragraphs describe two approaches, categorized based on modeling the covariance matrix $\bSigma$: diagonal and full covariance matrix.

\noindent\textbf{Diagonal Covariance Matrix.}
The most widely used way for uncertainty modeling is a diagonal covariance matrix.
It is parameter-efficient because the regressor only needs to output the mean and diagonal variance vector.
The diagonal Gaussian modeling needs twice the output dimensions, \ie, $g_{diag}: \Real^{d_{f}} \rightarrow \Real^{2 \cdot d_{o}}$ (See \Fref{fig:teaser}b).
The half dimension is for the mean vector, $\bmu \in \Real^{d_{o}}$, and the other half is for the variance vector, $\bsigma^2 \in \Real^{d_{o}}$ as follows:
\begin{equation}
    g_{diag}(f(\bx)) \sim \mathcal{N}(\bmu, \mathop{\mathrm{diag}}(\bsigma^2)).
\end{equation}
In implementation, instead of using direct $\bsigma^2$, the logarithm of $\bsigma^2$ is used with the exponential function mapping
% is applied to the variance vector 
to ensure the positive values.
The model is trained by minimizing the negative log-likelihood (NLL) loss with ground truth $\by$ as
\begin{equation}
\label{eq:diagnllloss}
    \calL_{NLL} = \log p(\by|\bmu, \bsigma^2) = \frac{||\by - \bmu || ^2}{2\bsigma^2} + 0.5 \log\bsigma^2.
\end{equation}

\begin{wraptable}{r}{0.48\linewidth}
\centering
    \resizebox{1.0\linewidth}{!}{
    \begin{tabular}{c c c c}
         \toprule
         Parametrization & Uncertainty modeling & $\#$ Params. & Example [$\#$ Params]\\
         \midrule
         Deterministic & \xmark &$d_fd_o$  & 0.065M\\
         Diagonal covariance & \cmark (independent)&$2d_fd_o$  & 0.129M\\
         Full covariance& \cmark & $d_f(d_o+d_o^2)$ & 4.129M \\
         Ours& \cmark & $2d_fd_o+d_o^2$ & 0.133M \\
         \bottomrule
    \end{tabular}}
    \vspace{4mm}
    \caption{\textbf{The number of parameters.} The dimensions of feature and output are $d_f$ and $d_o$, respectively. We list the required number of parameters for each parameterization. We set $d_f{=}1024, d_o{=}63$, and $k{=}21$ for example.}
    \label{tab:efficiency}
\end{wraptable} 

\noindent\textbf{Full Covariance Matrix.}
In a full covariance matrix, the regressor outputs all covariance matrix elements as
\begin{equation}
    g_{full}(f(\bx)) \sim \mathcal{N}(\bmu, \bSigma),
\end{equation}
where $\bmu \in \Real^{d_{o}}$ and $\bSigma \in \Real^{d_o \times d_o}$ (See \Fref{fig:teaser}c).
Since $\bSigma$ should be a positive definite matrix, we construct $\bSigma$ as $\bA\bA^\top$, which ensures $\bSigma$ to be positive definite.
Although it is less parameter-efficient compared to the diagonal covariance matrix, this approach has a higher capacity for capturing correlations.
The model is also trained by minimizing the NLL loss as follows:
\begin{equation}
\label{eq:fullnllloss}
    \calL_{NLL} = \log p(\by|\bmu, \bSigma) = \tfrac{1}{2}(\by - \bmu)^\top \bSigma^{-1} (\by - \bmu) + \tfrac{1}{2} \log\det\bSigma.
\end{equation}
The objective function involves the inverse covariance matrix.
% We observe that 
This introduces optimization instability.
Thus, it is more stable to directly estimate the precision matrix as $\bPsi=\bSigma^{-1}$.

In summary, the diagonal covariance matrix is parameter-efficient but does not capture the correlation; the full covariance matrix has opposite properties. 
We propose a new mid-representation that leverages a single linear layer which is parameter-efficient and effectively captures the hand joint correlation in a probabilistic manner.

\subsection{Correlation-Aware Aleatoric Uncertainty Estimation in Hand Joints}
Our goal is to estimate the aleatoric uncertainty of hand joints with the incorporation of joint correlation knowledge in an efficient yet expressive manner. The key idea is to define a probabilistic output space based on per-joint uncertainties learned under an independence assumption, and then transform it into a correlation-aware space using a single linear layer. 
%with Monte Carlo estimation. 
%We found that even a single linear layer can effectively and efficiently capture intrinsic hand joint dependencies.

\noindent\textbf{Aleatoric Uncertainty Estimation with Diagonal Covariance Matrix.}
We adapt HaMeR~\cite{pavlakos2024reconstructing}, a recently proposed transformer-based model for hand pose estimation pretrained on large-scale datasets to enable uncertainty modeling. In HaMeR, the Vision Transformer (ViT)~\cite{dosovitskiy2020image} backbone serves as a feature extractor and the mean 3D hand joint positions $\bmu_\mathbf{3D}$ are regressed by a transformer head followed by the MANO~\cite{romero2022embodied}. To enable uncertainty modeling, we incorporate an additional transformer head, denoted as uncertainty regressor, to regress the variance of each joint $\bsigma_\mathbf{3D}^2$. The model is trained by minimizing the 
% negative log-likelihood (
NLL loss (\Eref{eq:diagnllloss}) with ground truth 3D hand joint positions $\by_\mathbf{3D}$.
Up to here, the uncertainty modeling is same with the diagonal covariance matrix.

\begin{figure*}
\begin{center}
\includegraphics[width=\textwidth, keepaspectratio]{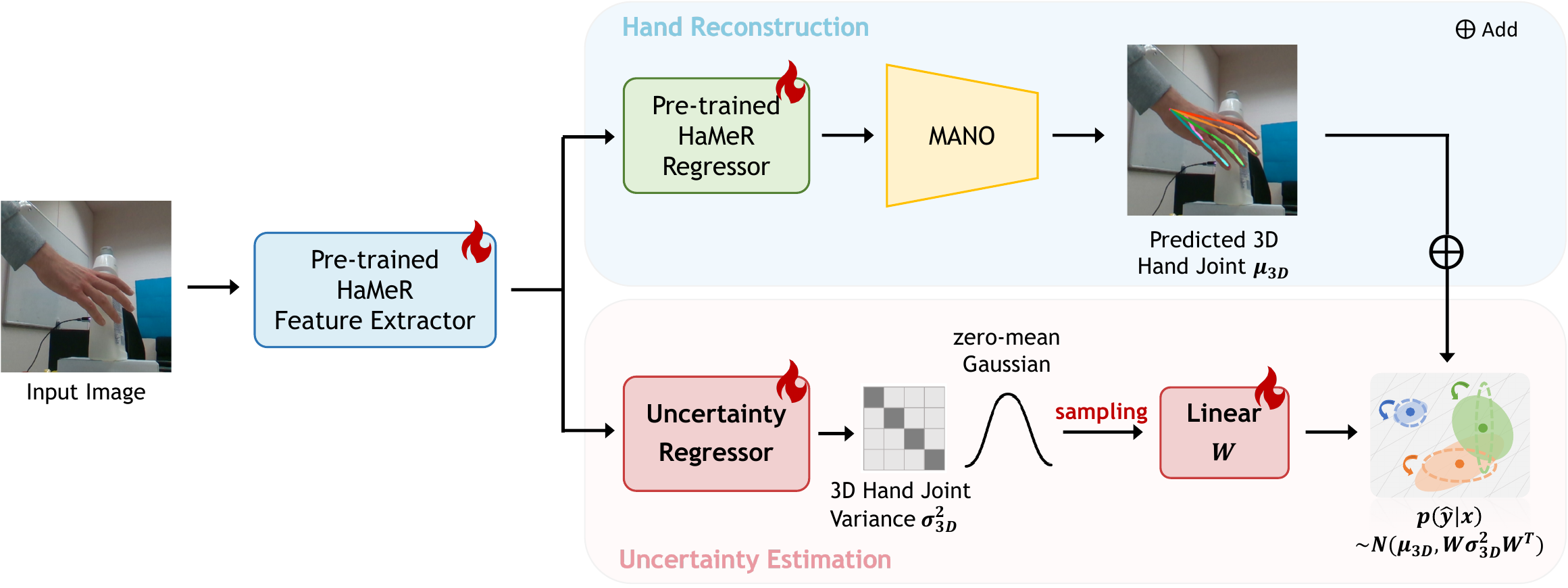}
\end{center}
   \caption{\textbf{Pipeline of our proposed method.} We train a pre-trained large model~\cite{pavlakos2024reconstructing} by introducing an additional transformer head that estimates the variance for each joint under the independent assumption. This estimated uncertainty then defines a probabilistic hand joint output space, from which we sample and pass the samples through a single linear layer to model the correlations between hand joints.}  
\label{fig:pipeline}
\end{figure*}

\noindent\textbf{Probabilistic Hand Joint Output Space.}
We define a probabilistic hand joint output space under a zero-mean Gaussian assumption, where per-joint variances learned under an independence assumption serve as the diagonal elements of the covariance matrix as $p(\bz|\bx) \sim \calN(\mathbf{0}, \mathop{\mathrm{diag}(\bsigma_\mathbf{3D}^2)})$. We draw $N$ samples from $p(\bz|\bx)$ and feed-forward these samples to single linear layer of weights $\bW$, ensuring that the final output follows $p(\hat{\mathbf{y}}|\bz) \sim \calN(\mathbf{0}, \bW\mathop{\mathrm{diag}}(\bsigma_\mathbf{3D}^2)\bW^\top)$ by the linearity of the Gaussian. By adopting the zero-mean assumption, we remove the dependency between the weights $\bW$ and the mean $\bmu_\mathbf{3D}$, enabling the model to focus solely on capturing correlations. Afterwards, the mean 3D hand joint positions $\bmu_\mathbf{3D}$ are added to this output space. As a result, we obtain:
\begin{equation}
\label{eq:fullnllloss}
    p(\hat{\mathbf{y}}|\bz) \sim \calN(\bmu_\mathbf{3D}, \bW\mathop{\mathrm{diag}}(\bsigma_\mathbf{3D}^2)\bW^\top).
\end{equation}
%After that, mean 3D hand joint positions $\bmu_\mathbf{3D}$ are are added to this output space to finally follows $p(\hat{\mathbf{y}}|\bz) \sim \calN(\bmu_\mathbf{3D}, \bW\mathop{\mathrm{diag}}(\bsigma^2_f)\bW^\top)$. 
We minimize the mean squared error (MSE) between the predictions and the ground truth:
\begin{equation}
\label{eq:mseloss}
    \calL_{MSE}=\mathbb{E}||\by - \hat{\mathbf{y}}||^2_2,
\end{equation}
where $\mathbb{E}$ is the expectation.
The overall training loss is defined as follows:
\begin{equation}
    \calL = \calL_{DETER}+ \lambda_{NLL}\calL_{NLL} + \lambda_{MSE}\calL_{MSE},
\end{equation}
where $\calL_{DETER}$ denotes the loss functions used for deterministic 3D hand pose estimation in HaMeR~\cite{pavlakos2024reconstructing}, and $\lambda_{NLL}$ and $\lambda_{MSE}$ are the weight factors for the uncertainty modeling terms. 

By sampling from the probabilistic hand joint output space and applying a single linear transformation, we naturally capture the correlation between hand joints, which can also be formulated analytically.
We position our method as a mid-representation between diagonal and full covariance modeling.
First, it offers higher expressiveness than diagonal covariance modeling in capturing inherent correlations between hand joints.
By adopting probabilistic space, our output distribution models hand joint correlation structure as full covariance modeling.
Second, our method requires less number of parameters compared to full covariance modeling, which involves the square of output dimension—making it impractical for high-dimensional outputs and prone to optimization instability.
In contrast, our approach introduces only a single linear transformation matrix on top of diagonal covariance modeling, providing both computational efficiency (see \Tref{tab:efficiency}) and improved optimization stability.
\section{Experiments}
\label{sec:experiments}
\subsection{Experimental Setup}
\noindent\textbf{Datasets.} 
We train our method on 2.7M training examples from multiple datasets that provide 2D or 3D hand annotations as in HaMeR~\cite{pavlakos2024reconstructing}. This includes FreiHAND~\cite{zimmermann2019freihand}, HO3D~\cite{hampali2020honnotate},
MTC~\cite{xiang2019monocular}, 
RHD~\cite{zimmermann2017learning}, InterHand2.6M~\cite{moon2020interhand2}, H2O3D~\cite{hampali2020honnotate},
DEX YCB~\cite{chao2021dexycb},
COCO WholeBody~\cite{jin2020whole},
Halpe~\cite{fang2022alphapose} and
MPII NZSL~\cite{simon2017hand}. 
To evaluate the quality of the estimated uncertainty and 3D hand pose estimation accuracy of our method, we use two standard benchmarks for 3D hand pose estimation, FreiHAND~\cite{zimmermann2019freihand} and HO3Dv2~\cite{hampali2020honnotate} which provide ground truth 3D hand annotations.  

\begin{wrapfigure}{r}{0.38\linewidth}
\centering
\vspace{-6mm}
    \resizebox{0.95\linewidth}{!}{
    \includegraphics[width=\textwidth, keepaspectratio]{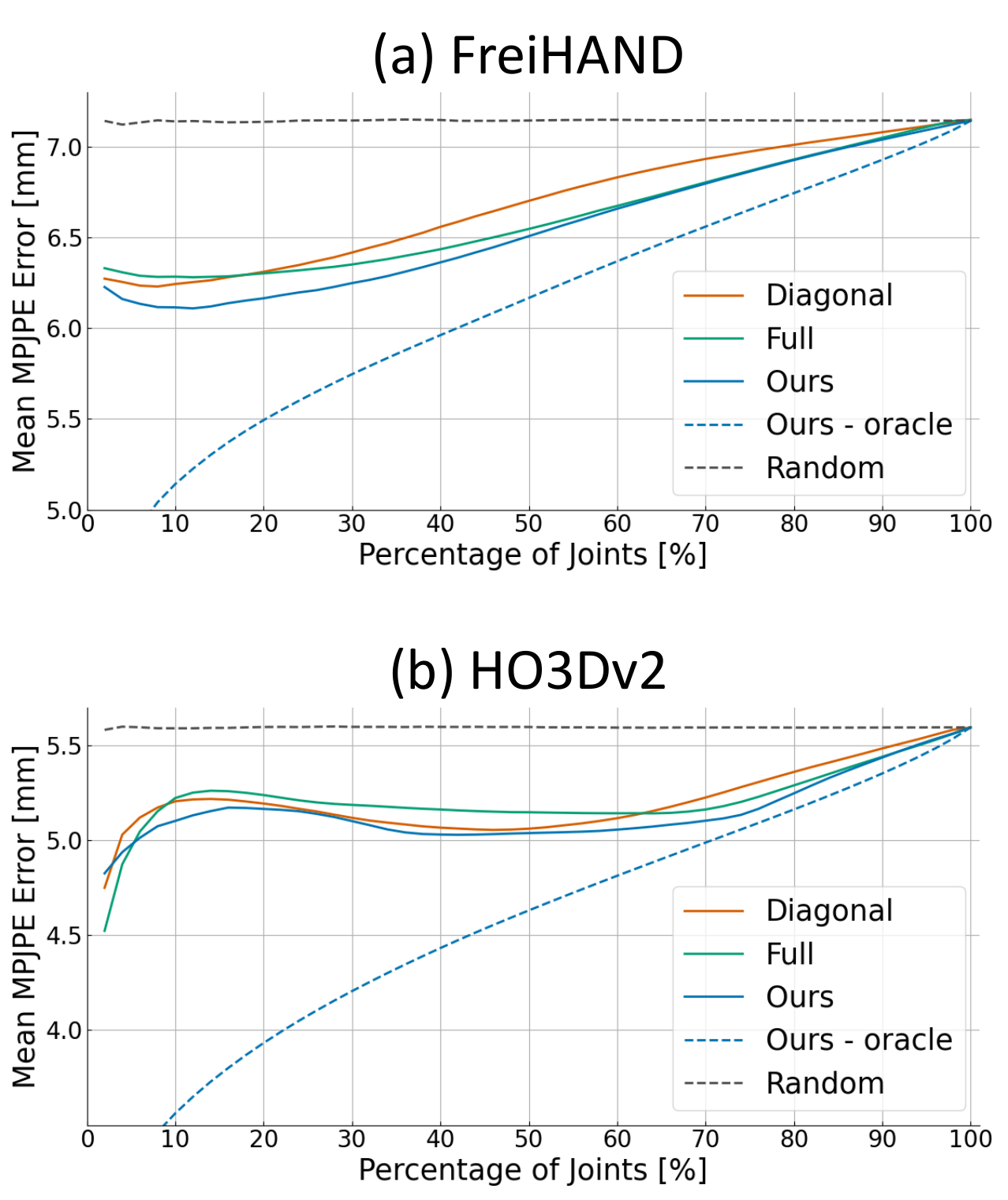}
    }
    \vspace{3mm}
    \caption{\textbf{Sparsification curves.} We compare sparsification curves obtained by different methods of estimating the uncertainty of 3D hand joints.}
    \label{fig:sparsification_curves}
\vspace{-2mm}
\end{wrapfigure} 

\noindent\textbf{Metrics on 3D hand pose estimation.}
We follow the typical protocols used in previous works~\cite{lin2021mesh,park2022handoccnet,pavlakos2024reconstructing}, and report PA-MPJPE and $\textrm{AUC}_\textrm{J}$ for evaluating estimated 3D hand joints and PA-MPVPE, $\textrm{AUC}_\textrm{V}$, F@5mm and F@15mm for evaluating estimated 3D hand mesh. PA-MPJPE and PA-MPVPE are measured in mm.

\noindent\textbf{Metrics on uncertainty estimation.}
To evaluate the quality of the estimated uncertainty, we use sparsification curves~\cite{poggi2020uncertainty, ilg2018uncertainty}. The predicted hand joints are sorted based on the estimated uncertainty. 
Given an error metric $\epsilon$, we evaluate the top $x$\% most certain joints. 
Ideally, if the uncertainty estimates are well-calibrated, the error is supposed to decrease as uncertain predictions are removed. We vary $x$ from 2 to 100, incrementing by 2, and report the area under the sparsification curve (AUSC) as in previous work~\cite{hu2012quantitative}. This metric is affected by both prediction accuracy and how similar the uncertainty-based sorting is to the actual error-based sorting. To only evaluate the latter, we also report the area under the sparsification error (AUSE)~\cite{ilg2018uncertainty} by subtracting the
oracle sparsification, which is obtained by sorting joints based on the $\epsilon$ magnitude, from
the estimated sparsification curve. We assume MPJPE as $\epsilon$. The uncertainty of each joint is measured as the trace of its estimated covariance matrix.
Additionally, we report Pearson's correlation coefficient, $\rho$, to quantify the degree to which the predicted uncertainty correlates with the true error. Pearson’s $\rho$ measures the strength and direction of the linear relationship between two continuous variables, providing a value between -1 and 1.

\begin{table}
\centering
\resizebox{0.85\linewidth}{!}{
\begin{tabular}{@{}lcccccc@{}}
\toprule
\multirow{2}{*}{Method} & \multicolumn{3}{c}{FreiHAND~\cite{zimmermann2019freihand}} & \multicolumn{3}{c}{HO3Dv2~\cite{hampali2020honnotate}} \\
\cmidrule(lr){2-4}\cmidrule(lr){5-7} \cmidrule(lr){5-5}
& AUSC $\downarrow$ & AUSE $\downarrow$ & Pearson's $\rho$ $\uparrow$ & AUSC $\downarrow$ & AUSE $\downarrow$ & Pearson's $\rho$ $\uparrow$\\
\midrule
\textit{Diagonal}&655&54.6&0.393& 511 & 63.3 & 0.448\\
\textit{Full}&648&47.6&0.453& 512 & 64.3 & 0.493\\
Ours&\textbf{642}&\textbf{42.2}&\textbf{0.569}& \textbf{505} & \textbf{57.6} & \textbf{0.600}\\
\bottomrule
\end{tabular}}
\vspace{4mm}
\caption{\textbf{Quantitative evaluation on uncertainty estimation.} We demonstrate the effectiveness of our parametrization in consistently enhancing all three metrics.}
\label{tab:quantitative_uncertainty}
\end{table}

\begin{figure}[t]
\begin{center}
\resizebox{1.0\linewidth}{!}{
    \includegraphics[width=\textwidth, keepaspectratio]{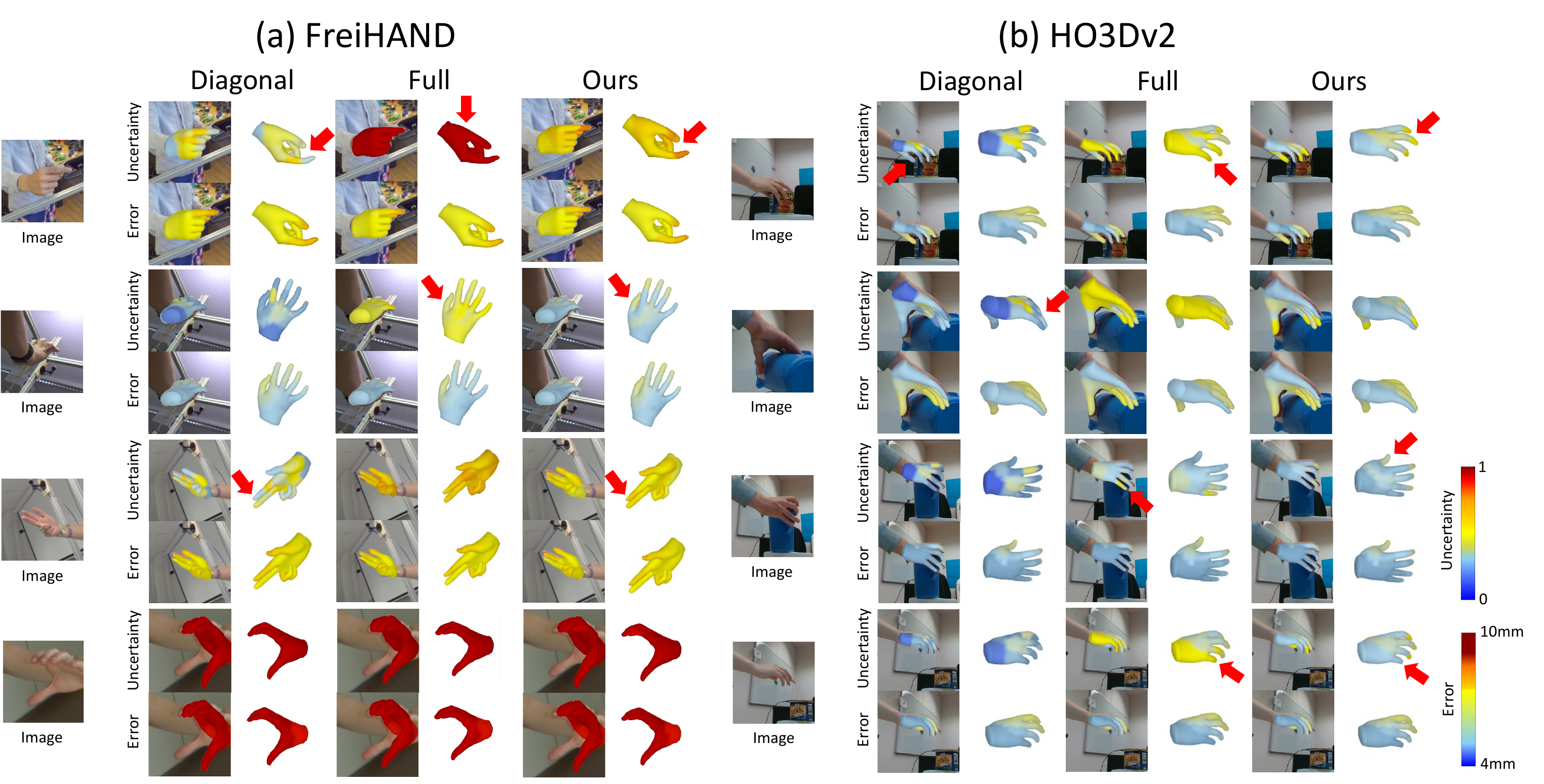}
    }
\end{center}
   \caption{\textbf{Qualitative results of uncertainty estimation.} We evaluate the quality of uncertainty estimates by comparing our method with existing uncertainty modeling methods. Specifically, we visualize the prediction errors alongside the corresponding uncertainty values. A desirable property of an uncertainty estimator is its proportionality to the actual prediction error—i.e., higher uncertainty values should correspond to higher prediction errors. The uncertainty estimated by our method shows stronger correlation with the prediction error, indicating its effectiveness in capturing model confidence.}  
\label{fig:qualitative}
\end{figure}

\noindent\textbf{Baselines.}
%As no previous work with publicly available code has estimated the 3D hand joint uncertainty, 
We implement two conventional uncertainty modeling methods based on Gaussian negative log-likelihood (NLL) as baselines and compare them with our method. \textit{(1) Diagonal:} The uncertainty regressor outputs per-joint variances under the independent assumption; the uncertainty is modeled using a diagonal covariance matrix. \textit{(2) Full:} The uncertainty regressor outputs all elements of the covariance matrix, modeling the uncertainty with a full covariance structure (see \Sref{sec:preliminary}). 

\noindent\textbf{Implementation details.}
In all experiments, including both the baselines and our method, we set $N$ = 25, $\lambda_{NLL}$ = 5$e$-4. We additionally set $\lambda_{MSE}$ = 5$e$-4 in our method. We train the models on a single NVIDIA A100 GPU for 550k iterations using the AdamW optimizer with a weight decay of 1$e$-4, $\beta_1$ = 0.9, and $\beta_2$ = 0.999. The learning rate is initialized as 1$e$-6, and
the mini-batch size is set to 64. All other experimental settings follow those used in HaMeR~\cite{pavlakos2024reconstructing}.

\subsection{Uncertainty Estimation}

\begin{figure*}
\begin{center}
\resizebox{0.85\linewidth}{!}{
\includegraphics[width=\textwidth, keepaspectratio]{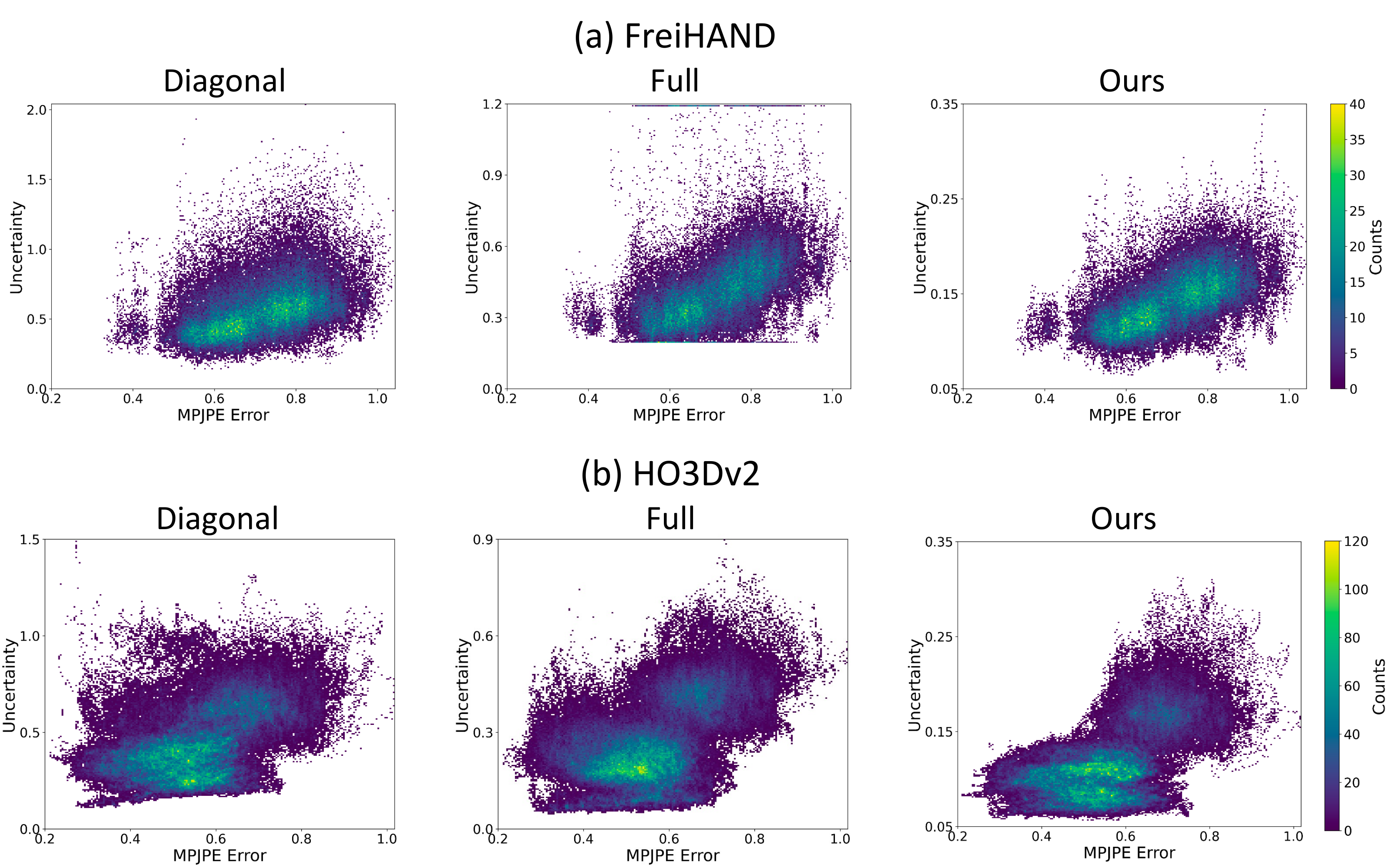}
}
\end{center}
   \caption{\textbf{2D histograms of model error and estimated uncertainty.} We show 2D histograms of the x-axis representing the model’s MPJPE error and the y-axis representing the estimated uncertainty on (a) FreiHAND~\cite{zimmermann2019freihand} and (b) HO3Dv2~\cite{hampali2020honnotate} datasets.}
\label{fig:histogram}
\end{figure*}

We conduct evaluations to assess the quality of the estimated uncertainty. The quantitative comparison with existing uncertainty modeling methods is presented in \Tref{tab:quantitative_uncertainty}. The results demonstrate that our method outperforms the baselines across all metrics. The sparsification curves
in \Fref{fig:sparsification_curves} show that although all methods perform similarly when evaluated on all 3D joints, our method achieves significantly higher accuracy than the others as 3D joints with high uncertainty are removed. This indicates that our uncertainty better reflects true 3D joint prediction errors. We additionally validate this by a qualitative comparision in \Fref{fig:qualitative}, wherein our uncertainty correlates better with the prediction error. Furthermore, we present qualitative results of the Pearson correlation by visualizing a 2D histogram of the model's MPJPE error and its estimated uncertainty. As shown in \Fref{fig:histogram}, the uncertainty estimates from our proposed method correlate with the model's error better than those from the \textit{Diagonal} and \textit{Full} baselines, while exhibiting fewer outliers.

\subsection{3D Hand Pose Estimation}

\begin{table}[t]
\centering
\resizebox{1.0\linewidth}{!}{
\begin{tabular}{@{}lcccccccccc@{}}
\toprule
\multirow{2}{*}{Method} & \multicolumn{4}{c}{FreiHAND~\cite{zimmermann2019freihand}} & \multicolumn{6}{c}{HO3Dv2~\cite{hampali2020honnotate}} \\
\cmidrule(lr){2-5}\cmidrule(lr){6-11}
& {PA-MPJPE $\downarrow$} & {PA-MPVPE $\downarrow$} & {F@5 $\uparrow$} & {F@15 $\uparrow$} & $\textrm{AUC}_\textrm{J}$ $\uparrow$ & PA-MPJPE $\downarrow$ & $\textrm{AUC}_\textrm{V}$ $\uparrow$ & PA-MPVPE $\downarrow$ & F@5 $\uparrow$ & F@15 $\uparrow$\\ 
\midrule
I2L-MeshNet~\cite{moon2020i2l} & 7.4 & 7.6 & 0.681 & 0.973 & 0.775 & 11.2 &0.722&13.9&0.409&0.932\\
Pose2Mesh~\cite{choi2020pose2mesh} & 7.7 & 7.8 & 0.674 & 0.969 & 0.754 & 12.5 &0.749&12.7&0.441&0.909\\
I2UV-HandNet~\cite{chen2021i2uv} & 6.7 & 6.9 & 0.707 & 0.977 & 0.804 & 9.9 &0.799&10.1&0.500&0.943 \\
METRO~\cite{lin2021end} & 6.5 & 6.3 & 0.731 & 0.984 & 0.792 & 10.4 & 0.779 & 11.1 & 0.484 & 0.946\\
MobRecon~\cite{chen2022mobrecon} & {\bf 5.7} & \underline{5.8} & \underline{0.784} & 0.986 & -&9.2&-& 9.4& 0.538& 0.957\\
AMVUR~\cite{jiang2023probabilistic} & 6.2 & 6.1 & 0.767 & 0.987 & 0.835 & 8.3 & 0.836 & 8.2 & 0.608 & 0.965\\ 
HaMeR~\cite{pavlakos2024reconstructing} & \underline{6.0} & {\bf 5.7} & {\bf 0.785} & {\bf 0.990} & 0.846 & \underline{7.7} & 0.841 & \underline{7.9} & 0.635 & \underline{0.980}\\ 
\midrule
\textit{Deterministic} & 6.1 & {\bf 5.7} & {\bf 0.785} & {\bf 0.990} & 0.845 & 7.8 & 0.840 & 8.0&0.629&\underline{0.980}\\ 
\textit{Diagonal} & 6.1 & \underline{5.8} & 0.782 & {\bf 0.990} & \underline{0.847} & \underline{7.7} & 0.842 & \underline{7.9}&0.638&{\bf 0.981}\\ 
\textit{Full} & 6.1 & 5.9 & 0.773 & \underline{0.989} & {\bf 0.849} & {\bf 7.6} & {\bf 0.844} & {\bf 7.8}&{\bf 0.644}&{\bf 0.981}\\ 
Ours & \underline{6.0} & {\bf 5.7} & \underline{0.784} & {\bf 0.990} & \underline{0.847} & {\bf 7.6} & \underline{0.843} & \underline{7.9}&\underline{0.640}&{\bf 0.981}\\ 
\bottomrule
\end{tabular}}
\vspace{4mm}
\caption{\textbf{Quantitative evaluation on 3D hand pose estimation.} Our method maintains comparable performance compared to 3D hand pose estimation methods and uncertainty modeling baselines.}
\label{tab:quantitative_hand_pose}
\end{table} 

\begin{table}
\centering
\resizebox{0.85\linewidth}{!}{
\begin{tabular}{@{}lcccccc@{}}
\toprule
\multirow{2}{*}{Method} & \multicolumn{3}{c}{FreiHAND~\cite{zimmermann2019freihand}} & \multicolumn{3}{c}{HO3Dv2~\cite{hampali2020honnotate}} \\
\cmidrule(lr){2-4}\cmidrule(lr){5-7} \cmidrule(lr){5-5}
& AUSC $\downarrow$ & AUSE $\downarrow$ & Pearson's $\rho$ $\uparrow$ & AUSC $\downarrow$ & AUSE $\downarrow$ & Pearson's $\rho$ $\uparrow$\\
\midrule
Ours w/o linear layer &648&47.9&0.544&509&61.4&0.524\\
Ours&\textbf{642}&\textbf{42.2}&\textbf{0.569}& \textbf{505} & \textbf{57.6} & \textbf{0.600}\\
\bottomrule
\end{tabular}}
\vspace{4mm}
\caption{\textbf{Ablation study on joint correlation in uncertainty estimation.} We show that our joint correlation modeling consistently improves uncertainty estimation performance across all three metrics.}
\label{tab:quantitative_uncertainty_ablation}
\end{table}

\begin{table}[h!]
\centering
\resizebox{1.0\linewidth}{!}{
\begin{tabular}{@{}lcccccccccc@{}}
\toprule
\multirow{2}{*}{Method} & \multicolumn{4}{c}{FreiHAND~\cite{zimmermann2019freihand}} & \multicolumn{6}{c}{HO3Dv2~\cite{hampali2020honnotate}} \\
\cmidrule(lr){2-5}\cmidrule(lr){6-11}
& {PA-MPJPE $\downarrow$} & {PA-MPVPE $\downarrow$} & {F@5 $\uparrow$} & {F@15 $\uparrow$} & $\textrm{AUC}_\textrm{J}$ $\uparrow$ & PA-MPJPE $\downarrow$ & $\textrm{AUC}_\textrm{V}$ $\uparrow$ & PA-MPVPE $\downarrow$ & F@5 $\uparrow$ & F@15 $\uparrow$\\ 
\midrule
Ours w/o linear layer & 6.1 & 5.8 & 0.782 & {\bf 0.990} & {\bf 0.847} &  7.7 & {\bf 0.843} & {\bf 7.9}&0.638&{\bf 0.981}\\ 
Ours & {\bf6.0} & {\bf 5.7} & {\bf 0.784} & {\bf 0.990} & {\bf0.847} & {\bf 7.6} & {\bf 0.843} & {\bf 7.9}&{\bf0.640}&{\bf 0.981}\\ 
\bottomrule
\end{tabular}}
\vspace{4mm}
\caption{\textbf{Ablation study on joint correlation in 3D hand pose estimation.} We demonstrate that our joint correlation modeling shows improvement in 3D hand pose estimation accuracy.}
\label{tab:quantitative_hand_pose_ablation}
\end{table}

We evaluate the 3D hand pose estimation capability of our method. 
\Tref{tab:quantitative_hand_pose} compares our approach with existing 3D hand pose estimation methods that do not support uncertainty estimation, as well as with uncertainty modeling baselines. For a fair comparison, we also assess our method against a deterministic baseline denoted as \textit{Deterministic}, which fine-tunes the pre-trained network without incorporating uncertainty modeling. Compared to 3D hand pose estimation methods and deterministic baseline, our model achieves competitive performance on both benchmarks, while additionally providing uncertainty estimation. In the comparison of uncertainty modeling baselines, although the full baseline achieves slightly better 3D hand pose estimation performance on the HO3Dv2~\cite{hampali2020honnotate} evaluation dataset, our method offers far
% significantly
more reliable uncertainty estimates through efficient joint correlation modeling (see Tables~\ref{tab:efficiency} \& \ref{tab:quantitative_uncertainty}).

\subsection{Ablation Study}
\noindent\textbf{Ablation study on joint correlation.}
We conduct an ablation study to investigate whether modeling joint correlations using a linear layer improves uncertainty estimation. Specifically, we remove the linear layer, resulting in a probabilistic hand joint output space constructed under an independence assumption, where uncertainty is modeled with per-joint variance. The quantitative results in Tables~\ref{tab:quantitative_uncertainty_ablation} and \ref{tab:quantitative_hand_pose_ablation} support our conclusion that incorporating the linear layer to capture joint correlations enhances the performance of uncertainty estimation and 3D hand pose estimation.

\begin{figure*}
  \centering
  \begin{minipage}{0.46\textwidth}
    \centering
    \resizebox{0.8\linewidth}{!}{
    \includegraphics[width=\textwidth, keepaspectratio]{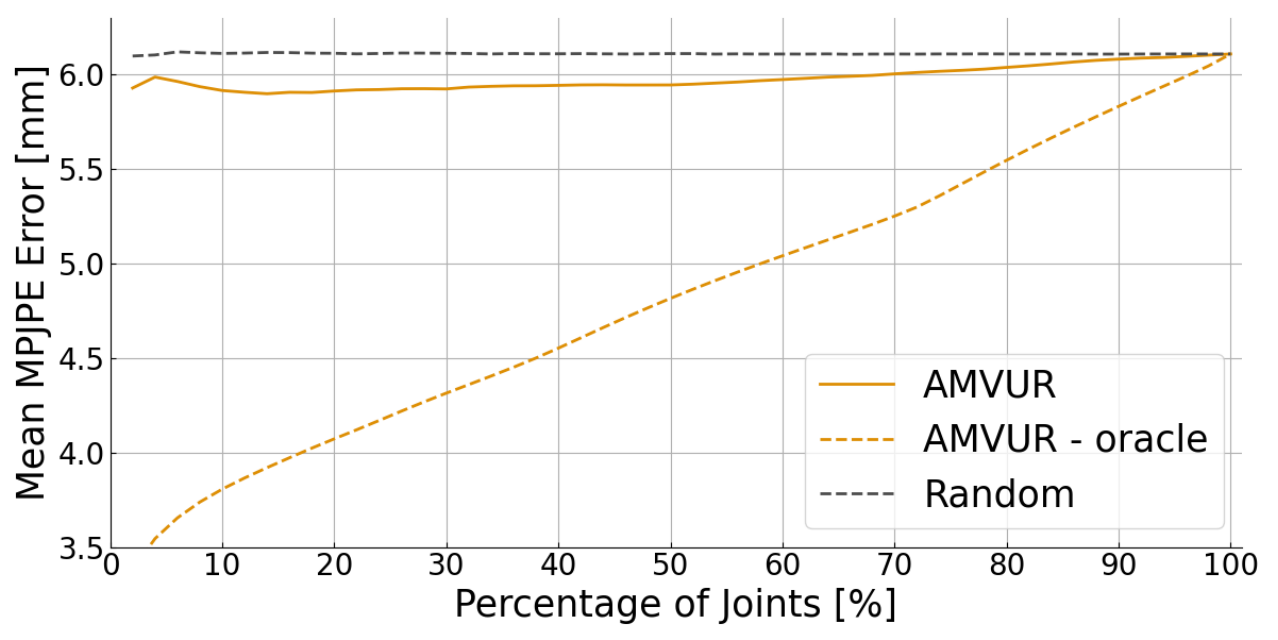}
    }
    \vspace{2mm}
    \caption{\textbf{Sparsification curve of AMVUR~\cite{jiang2023probabilistic}.} The competing method exhibits a large discrepancy between uncertainty-based and error-based sortings.}
    \label{fig:sparsification_curve_amvur}
  \end{minipage}
  \hfill
  \begin{minipage}{0.5\textwidth}
    \centering
    \resizebox{0.95\linewidth}{!}{
    \begin{tabular}{@{}lccc@{}}
    \toprule
    Method&AUSC $\downarrow$ & AUSE $\downarrow$& Pearson's $\rho$ $\uparrow$ \\
    \midrule
    AMVUR~\cite{jiang2023probabilistic}&586&113&0.127\\
    Ours&\textbf{505} & \textbf{57.6} & \textbf{0.600}\\
    \bottomrule
    \end{tabular}}
    \vspace{3mm}
    \captionof{table}{\textbf{Quantitative comparison with AMVUR~\cite{jiang2023probabilistic}.} Our method demonstrates consistently better performance than the competing method across all uncertainty evaluation metrics.}
    \label{tab:quantitative_amvur}
  \end{minipage}
\end{figure*}

\begin{table}[h!]
\centering
\resizebox{0.85\linewidth}{!}{
\begin{tabular}{ccccccc}
\toprule
\multirow{2}{*}{$\#$ Samples} & \multicolumn{3}{c}{FreiHAND~\cite{zimmermann2019freihand}} & \multicolumn{3}{c}{HO3Dv2~\cite{hampali2020honnotate}} \\
\cmidrule(lr){2-4}\cmidrule(lr){5-7} \cmidrule(lr){5-5}
& AUSC $\downarrow$ & AUSE $\downarrow$ & Pearson's $\rho$ $\uparrow$ & AUSC $\downarrow$ & AUSE $\downarrow$ & Pearson's $\rho$ $\uparrow$\\
\midrule
1&653&52.7&0.399& 507 & 58.6 & 0.557\\
5&655&54.3&0.414& 510 & 62.4 & 0.484\\
10&\underline{648}&\underline{47.1}&\underline{0.551}& \underline{507} & \underline{59.1} & \underline{0.536}\\
25&\textbf{642}&\textbf{42.2}&\textbf{0.569}& \textbf{505} & \textbf{57.6} & \textbf{0.600}\\
%50&\underline{646}&\underline{45.3}&0.544& 508 & 61.1 & 0.509\\
%100&648&48.0&0.535& - & - & -\\
\bottomrule
\end{tabular}}
\vspace{4mm}
\caption{\textbf{Effect of the number of samples on uncertainty estimation.} We change the number of samples in output space from 1 to 25 and evaluate the quality of estimated uncertainty. }
\label{tab:num_of_samples}
\end{table}

\noindent\textbf{Comparison with related work.}
We compare the uncertainty estimation quality of our method on the HO3Dv2~\cite{hampali2020honnotate} dataset with an open-sourced competing method~\cite{jiang2023probabilistic}, which models the 3D hand joint output space as a probabilistic distribution and inherently supports uncertainty measurement. As shown in \Tref{tab:quantitative_amvur}, our method outperforms the competing approach in terms of uncertainty estimation performance. Furthermore, the sparsification curve of AMVUR~\cite{jiang2023probabilistic} in \Fref{fig:sparsification_curve_amvur} indicates that it struggles to align uncertainty-based sorting with actual error-based sorting.

\noindent\textbf{Ablation study on number of samples.}
Our method draws the samples in the output space and feed-forward these samples to the regressor. We conduct an ablation study to investigate the effect of the number of samples on uncertainty estimation by varying this number. \Tref{tab:num_of_samples} presents the results. 
In general, the larger the number of samples, the better the performance.
We select the default number as 25, considering the trade-off between performance and computational complexity.

\section{Conclusion}
\label{sec:conclusion}
This paper addresses two key challenges in existing 3D hand pose estimation models: the inability to estimate aleatoric uncertainty, and the lack of uncertainty modeling that incorporates joint correlation knowledge. We estimate and evaluate the aleatoric uncertainty in 3D hand pose estimation. We propose a new parametrization that leverages a single linear layer, which effectively captures the inherent hand joint correlation and achieves a better trade-off between modeling joint correlations and computational efficiency. To enable this, we formulate a probabilistic hand joint output space and then transform it into the correlation-aware space using the single linear layer. Experimental results show that the estimated aleatoric uncertainty of our method correlates well with the prediction error while maintaining 3D hand pose estimation performance.
%-------------------------------------------------------------------------
\clearpage
\paragraph{Acknowledgments}
This research was supported by Institute of Information \& communications Technology Planning \& Evaluation (IITP) grant funded by the Korea government(MSIT) (No.RS-2025-25443318, Physically-grounded Intelligence: A Dual Competency Approach to Embodied AGI through Constructing and Reasoning in the Real World; No.RS-2019-II191906, Artificial Intelligence Graduate School Program(POSTECH)), and Culture, Sports and Tourism R\&D Program through the Korea Creative Content Agency grant funded by the Ministry of Culture, Sports and Tourism in 2024 (Project Name: Development of barrier-free experiential XR contents technology to improve accessibility to online activities for the physically disabled, Project Number: RS-2024-00396700, Contribution Rate: 20\%). It was also supported by the KAIST Cross-Generation Collaborative Lab Project, and the ‘Ministry of Science and ICT’ and NIPA (``HPC Support'' Project).

\bibliography{egbib}
\end{document}